\documentclass[10pt,twocolumn,letterpaper]{article}

\usepackage[pagenumbers]{cvpr} 

\usepackage{multirow}
\usepackage{algpseudocode}
\usepackage{graphicx}
\usepackage{algorithm}
\usepackage{relsize}
\usepackage{booktabs}
\usepackage{wrapfig}
\usepackage{multirow}
\usepackage{amsmath}
\usepackage{enumitem}
\usepackage{mathtools}
\usepackage{rotating}

\definecolor{cvprblue}{rgb}{0.21,0.49,0.74}
\usepackage[pagebackref,breaklinks,colorlinks,allcolors=cvprblue]{hyperref}

\def\confName{CVPR}
\def\confYear{2025}

\title{TPIE: Topology-Preserved Image Editing With Text Instructions}

\author{$\text{Nivetha Jayakumar}^1 $
\and
$\text{Srivardhan Reddy Gadila}^2$
\and
$\text{Tonmoy Hossain}^2$
\and
$\text{Yangfeng Ji}^2$
\and
$\text{Miaomiao Zhang}^{1,2}$
\\
$\text{Department of Electrical and Computer Engineering}^1$ \\
$\text{Department of Computer Science}^2$ \\
University of Virginia}

\begin{document}

\twocolumn[{%
\renewcommand\twocolumn[1][]{#1}
\maketitle
\begin{center}
    \captionsetup{type=figure, width=0.95\textwidth}
    \vspace{-4mm}
    \includegraphics[width=0.95\textwidth, trim = 0cm 0.2cm 0cm 0cm]{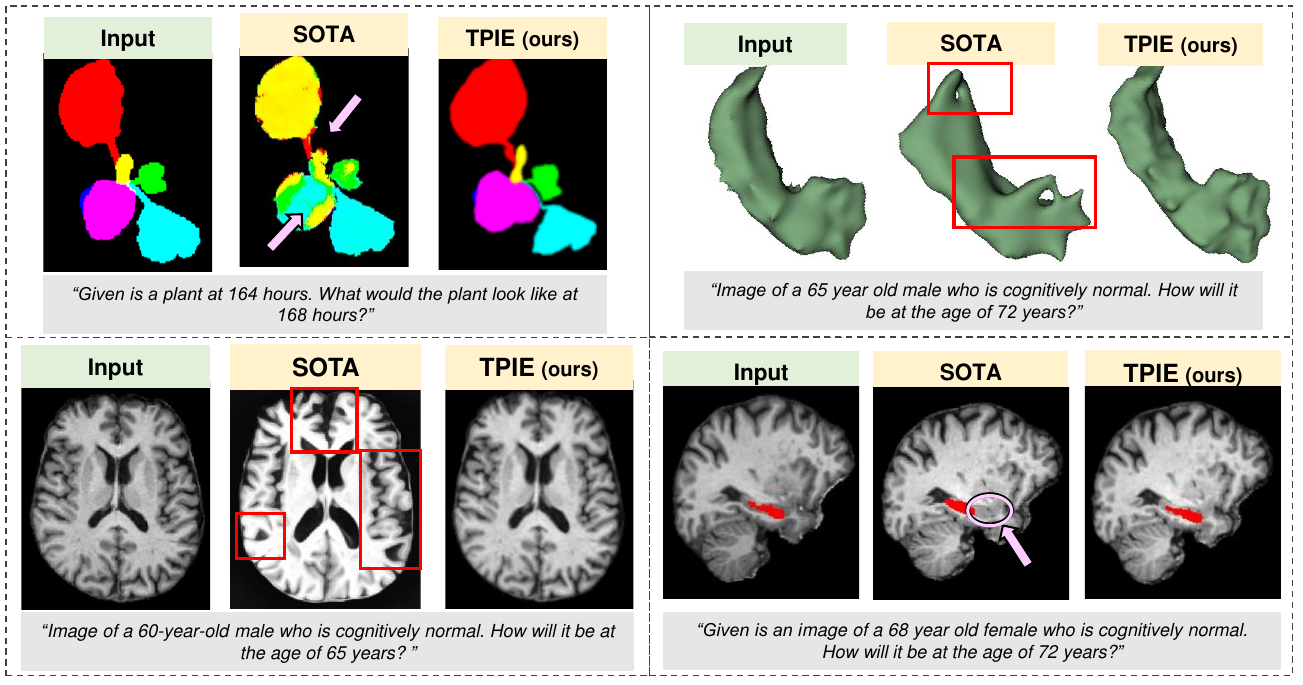}
    \captionof{figure}{An illustration of image samples generated by a state-of-the-art (SOTA) model - InstructPix2Pix (IP2P)~\cite{brooks2023instructpix2pix}, and our proposed model TPIE. The samples generated by IP2P  present challenges in preserving the object topology of individual structures. Violations of topology are highlighted by pink arrows and red boxes.}
    \label{fig:intro_graph}
\end{center}
\vspace{-0.9mm}
}]


\begin{abstract}

Preserving topological structures is important in real-world applications, particularly in sensitive domains such as healthcare and medicine, where the correctness of human anatomy is critical. However, most existing image editing models focus on manipulating intensity and texture features, often overlooking object geometry within images. To address this issue, this paper introduces a novel method, Topology-Preserved Image Editing with text instructions (TPIE), that for the first time ensures the topology and geometry remaining intact in edited images through text-guided generative diffusion models. More specifically, our method treats newly generated samples as deformable variations of a given input template, allowing for controllable and structure-preserving edits. Our proposed TPIE framework consists of two key modules: (i) an autoencoder-based registration network that learns latent representations of object transformations, parameterized by velocity fields, from pairwise training images; and (ii) a novel latent conditional geometric diffusion (LCDG) model efficiently capturing the data distribution of learned transformation features conditioned on custom-defined text instructions. We validate TPIE on a diverse set of 2D and 3D images and compare them with state-of-the-art image editing approaches. Experimental results show that our method outperforms other baselines in generating more realistic images with well-preserved topology. Our code will be made publicly available on Github.

\end{abstract}
 
\section{Introduction}
\label{sec:intro}

Generative diffusion models have gained increasing popularity over the past years for achieving groundbreaking performance in high-quality image editing and generation tasks~\cite{brock2019large,karras2020analyzing,mukhopadhyay2023diffusion}. Among these, language-image generative models \cite{ramesh2022hierarchical,saharia2022photorealistic,yu2022scaling} have been extensively explored due to its highly customizable and context-aware editing capabilities. In particular, these models leverage text prompts to guide the diffusion process, allowing users to make specific changes to images in the original content; hence producing diverse and contextually accurate visual modifications.

Text-guided diffusion models have been explored in a wide variety of fields, including content creation~\cite{chen2023fantasia3d, lin2023magic3d, huang2023dreamtime}, augmented reality~\cite{zhao2024clear, hossieni2024puzzlefusion}, and medical imaging~\cite{dorjsembe2023conditional, wu2024medsegdiff}. These models typically utilize text encoders (i.e., CLIP embeddings~\cite{radford2021learning}) with diffusion processes to align text with visual features effectively. 

While early models~\cite{nichol2022glide,ramesh2021zero} used simple text prompts to describe image semantics, more recent approaches~\cite{hertz2022prompttoprompt} leverage more complex, user-provided natural language instructions for editing. These models carry out fine-grained image synthesis by either providing representations of the template as image guidance ~\cite{brooks2023instructpix2pix, meng2021sdedit}, fine-tuning existing conditional models on a single image~\cite{valevski2023unitune, kawar2023imagic}, or incorporating additional semantic label-maps~\cite{nichol2022glide, wang2023instructedit}. In specialized domains, recent work~\cite{gu2023biomedjourney} has adapted these methods for counterfactual medical image generation by using a pre-trained IP2P network~\cite{hertz2022prompttoprompt}, showing the potential of text-guided diffusion in medical applications.

Despite the success of aforementioned approaches, existing models heavily manipulate image intensity and texture features; hence neglecting object topology and geometry in their editing or generation processes~\cite{nguyen2024visual,li2024blip,hertz2022prompttoprompt,valevski2023unitune, kawar2023imagic}. This often results in unrealistic image samples that pose risks to downstream tasks relying on generated image data~\cite{azizi2023synthetic}. The preservation of object topological structure is particularly critical in high stakes applications, such as healthcare and biomedical domains~\cite{jayakumar2023sadir,hu2019topology,santhirasekaram2023topology}. Recent generative models have started to introduce simple geometric constraints in the diffusion process, using various methods such as shape localization~\cite{patashnik2023localizing, gupta2024topodiffusionnet}, object boundary condition~\cite{maze2023diffusion}, and prior distributions of 3D shape representations in the form of meshes and point clouds~\cite{yu2025surf, hu2024topology}. However, none of these approaches have incorporated highly-detailed and fine-grained structural-preserving properties into image-based latent diffusion models while relying solely on text instructions. 

In this paper, we propose a novel text-guided image editing model, TPIE, that for the first time learns object geometry presented in images and preserves its topology based on latent diffusion models~\cite{rombach2022highresolution}. In contrast to current methods mainly manipulating image intensity and texture features~\cite{brooks2023instructpix2pix, meng2021sdedit}, our method considers generated samples as deformable variants of the given input with the same topology (as shown in Fig.~\ref{fig:intro_graph}). That is to say, each synthesized image will be generated by deforming the input template/reference image with a learned diffeomorphic transformation (a.k.a., one-to-one smooth and inverse smooth mapping) conditional on the text input. Our proposed TPIE includes two main modules: (i) an autoencoder-based registration network that learns latent representations of diffeomorphic transformations from pairwise training images; and (ii) a new latent conditional geometric diffusion model that efficiently captures the data distribution of latent transformation features conditioned on user-crafted text instructions. The main contributions of TPIE are summarized as:   

\begin{enumerate}[label=(\roman*)]
\item Formulating and incorporating geometric properties of image objects in text-guided diffusion models; 
\item Developing a new image editing model that, for the first time, learns latent distribution of geometric transformations aligned with text features;

\item Our work demonstrates great potential for controllable and topology-aware editing models, addressing the unique challenge of maintaining realistic anatomy in sensitive domains.
\end{enumerate}
We validate the performance of TPIE on a diverse set of real-world datasets, including 2D plant-growth data~\cite{Uchiyama2017komatsuna}, and real 3D brain MRIs and hippocampus shape data~\cite{LaMontagne2019oasis3}. We compare the model performance with state-of-the-art generative models used for image editing with text instructions. Experimental results show that TPIE achieves significantly improved results in generating image samples with well preserved topological structures.
\section{Background: Preserving Topology via Diffeomorphic Transformations}
\label{sec:background}
In this section, we briefly review the formulation of topological constraints through diffeomorphic transformations deforming one image into another~\cite{miller2004computational,grenander1998computational}. With the underlying assumption that objects of a generic class can be described as deformed versions of the others, descriptors of that class arise naturally by transforming/deforming a template image to all the other images in that class~\citep{avants2008symmetric,reuter2012within,joshi2004unbiased}. The resulting transformation is then considered as a representation that reflects geometric object changes. 

\begin{figure}[h]
\centering
\centerline{\includegraphics[width=\linewidth, trim = 0cm 0.5cm 0cm 0cm]{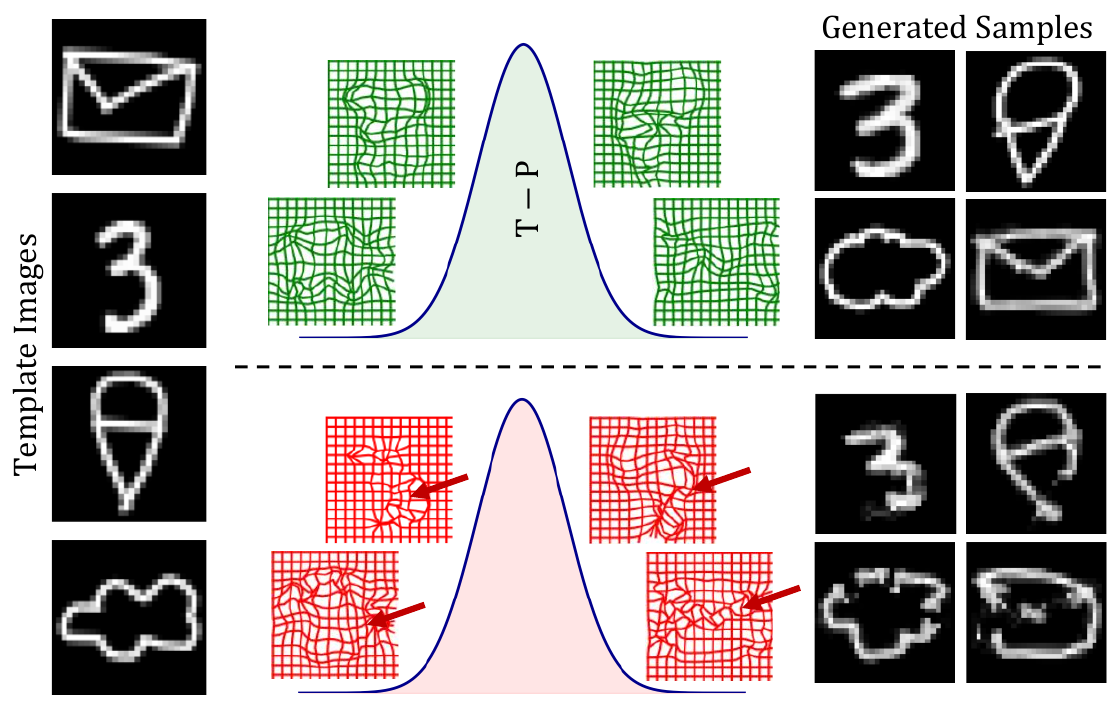}}
\caption{Examples of generated images deformed with sampled transformations from topology-preserved (T-P) distributions (top panel) vs not preserved (bottom panel).}
\label{fig:topology}
\vspace{-3mm}
\end{figure}

In theory, every topological property of the deformed template can be preserved by enforcing the transformation field to be diffeomorphisms, i.e., differentiable, bijective mappings with differentiable inverses~\citep{beg2005computing,arnold1966geometrie,miller2006geodesic}. Examples of generated images with vs. without well-preserved topology are shown in Fig.~\ref{fig:topology}. Violations of topological constraints in the deformation space introduce artifacts, such as tearing, crossing, or passing through itself (see pointed arrows in the deformation fields in Fig.~\ref{fig:topology}).\\

\noindent {\bf Parameterizing diffeomorphisms in the space of velocity fields.} Given a template image $S$ and a target image $T$ defined on a $d$-dimensional torus domain $\Omega = \mathbb{R}^d / \mathbb{Z}^d$ ($S(x), T(x):x \in \Omega \rightarrow \mathbb{R}$), a diffeomorphic transformation, $\phi_t$, for $t \in [0, 1]$, is defined as a smooth flow over time to deform a template image to a target image by a composite function, $S \circ \phi^{-1}_t$. Here, the $\circ$ denotes an interpolation operator. Such a transformation is typically parameterized by time-dependent velocity fields under a large diffeomorphic deformation metric~\cite{beg2005computing}, or a stationary velocity field (SVF) that remains constant over time~\cite{arsigny2006log}. While we employ SVF in this paper, our framework is easily applicable to the other. 

For a stationary velocity field $v$, the diffeomorphisms, $\phi_t$, are generated as solutions to the equation:
\begin{equation}
\label{eq:svf}
\frac{d\phi_t}{dt} = v \circ \phi_t,   \,\, \text{s.t.}  \,\, \,\, \phi_0 = x.
\end{equation}
The solution of Eq.~\eqref{eq:svf} is identified as a group exponential map using a scaling and squaring scheme~\cite{arsigny2006log}. More details are included in~\cite{arsigny2006log}. \\
\noindent {\bf Deriving diffeomorphisms from pairwise images.} The diffeomorphic transformation, $\phi_t$, that reflects geometric changes between images can be solved by minimizing an explicit energy functional as
\begin{equation}
\label{eq:tenergy}
E(v) = \eta \text{Dist}(S \circ \phi_1^{-1} (v), T) + \text{Reg}(v), \text{s.t. Eq.}~\eqref{eq:svf}. 
\end{equation}
Here Dist(·,·) is a  distance function that measures the dissimilarity between images, Reg($\cdot$) is a regularization term that enforces the smoothness of transformation fields, and $\eta$ is a positive weighting parameter. Widely used distance functions include the sum-of-squared intensity differences ($L_2$-norm)~\cite{beg2005computing}, normalized cross correlation (NCC)~\cite{avants2008symmetric}, and mutual information (MI)~\cite{wells1996multi}. In this paper, we utilize a sum-of-squared distance function. 
\section{Method: TPIE}

This section presents a novel framework, TPIE, that incorporates geometric information of objects in synthesized images using generative diffusion models with textual instructions. There are two main components developed in the TPIE framework: (i) an autoencoder-based deformable registration network that learns latent representations of diffeomorphic transformations, parameterized by velocity fields, from training images; and (ii) a novel latent conditional geometric diffusion model, LCDG, that efficiently captures the latent distribution of transformation features conditioned on texts. A joint loss function of both subnetworks will be miminized to optimal. Details of our network architecture are introduced as follows. 

\subsection{Representation learning of velocity fields}
For a number of $N$ template and target images associated with text instructions $\{m_i, f_i, \Lambda_i\}_{i=1}^N$, we employ an autoencoder-based deformable registration network~\cite{hinkle2018lagomorph, hinkle2018diffeomorphic} to learn the latent representation of velocity fields associated with pairwise input images (i.e., a template, $m_i$, and a target, $f_i$). The encoder of this network, parameterized by $\mathbf{E_\omega}$, produces the latent representation of the velocity field, denoted as $\{\boldsymbol{\gamma}_i\}^{C \times H \times W}$, where $C, H, W$ indicate the latent dimensions. The decoder, $\mathbf{D_\omega}$, then projects this latent velocity back to the high-dimensional input space, represented as $v_i\in \mathrm{V}$. The predicted velocity field, after being scaled and integrated, is passed to the spatial transform layer~\cite{jaderberg2015spatial} to deform the template image using Eq.~\eqref{eq:svf}; thereby generating the predicted target.

We formulate the registration loss that minimizes the dissimilarity between the ground truth target and the deformed template as:
\setlength{\abovedisplayskip}{1pt}
\begin{equation} 
\label{eq:registration}
\mathcal{L_\omega} = \frac{1}{N} \mathlarger{\sum}_{i=1}^N \sigma \| m_i \circ \phi_i^{-1}(v_i(\omega)) - f_i \|^2_2 + \| \nabla v_i\|^2_2 + \text{reg}(\omega),
\end{equation}
where $\text{reg}(\cdot)$ is a regularization term on network parameters.

\begin{figure*}[!htbp]
\centering
\centerline{\includegraphics[width=\textwidth, trim = 0cm 0.5cm 0cm 0cm]{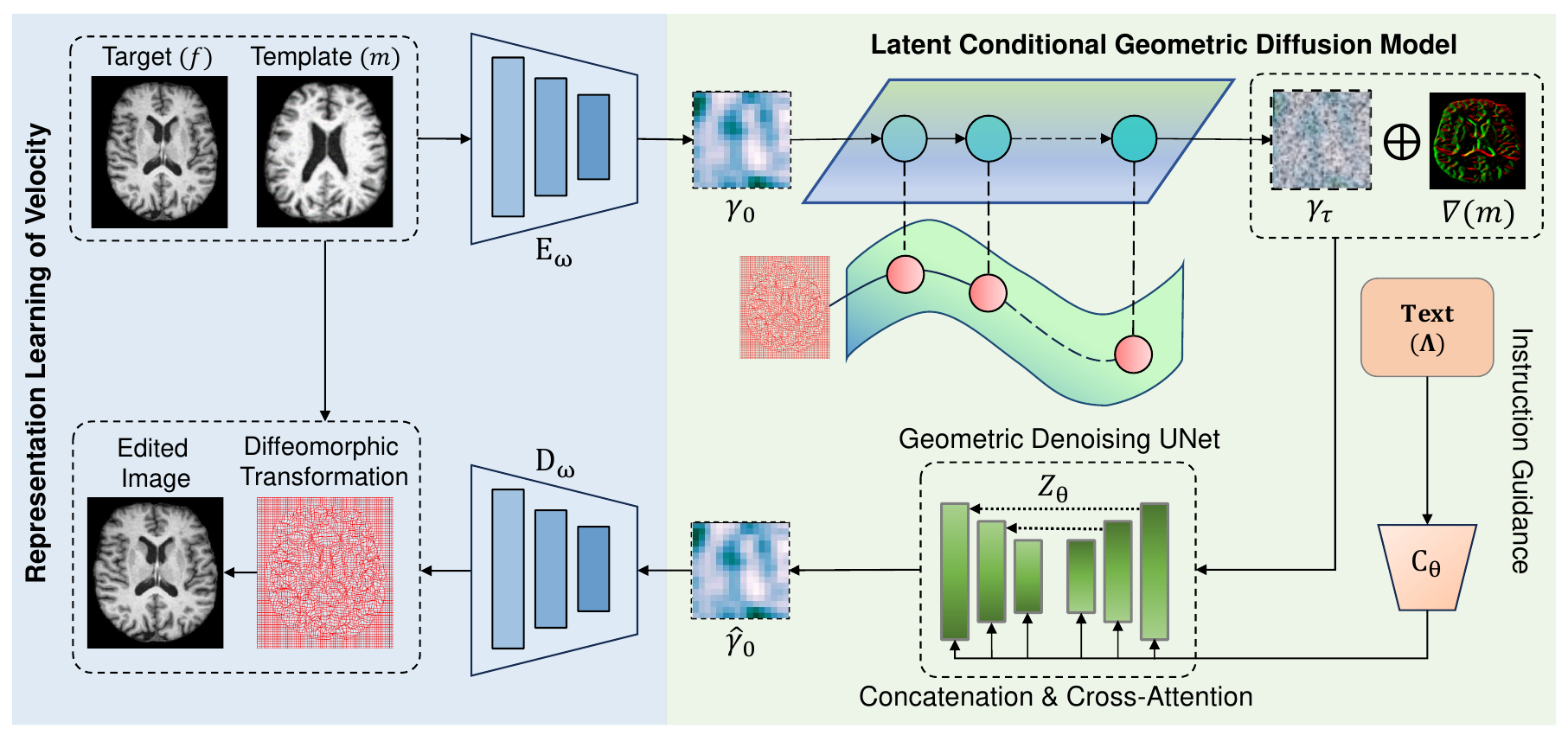}}
\caption{An overview of our proposed TPIE framework. There are two key components in the presented TPIE: (i) a latent representation learning of velocity network; and (ii) a latent conditional geometric diffusion (LCGD) model guided by text instructions.}
\label{fig:model}
\end{figure*}

\subsection{Latent conditional geometric diffusion model} 
In contrast to previous works that apply the diffusion process in the image space~\cite{brooks2023instructpix2pix,kim2022diffusemorph}, we are the first to develop a latent conditional diffusion model in the geometric deformation space. This includes (i) a scaling module that maps the input latent features of velocity fields, $v$, to a stable numerical range, ensuring computational stability; and (ii) a diffusion process directly sampling topology-preserved diffeomorphic transformations, which are then used to deform the input template to produce the final edited images. \\

\noindent {\bf \em - Scaling module in latent velocity space.} Since the velocity fields can take on large negative and positive values due to the nature of the underlying transformations, we introduce a scaling module in the latent geometric space to normalize the latent features to $[-1, 1]$. This helps prevent gradient explosion or vanishing issues and differs slightly from the original approach in the denoising diffusion probabilistic model~\cite{ho2020denoising}.

Given a set of $N$ latent features $\{\boldsymbol{\gamma}_{i}\}_{i=1}^N$, where $\boldsymbol{\gamma}_{i} \in \mathbb{R}^D$ with $D = C \times H \times W$. We define the scaling module, $\psi$, as: 
\begin{equation}
\begin{split}
    \psi({\boldsymbol{\gamma}_i}) = \frac{{\boldsymbol{\gamma_i}} - \boldsymbol{\Bar{\gamma}_i}}{\sqrt{\frac{\sum_{d=1}^{D} (\boldsymbol{\gamma_{i}^d} - \boldsymbol{\Bar{\gamma}_{i}})^2}{{D}}}},  \\
    \text{where}\,\,\, {\boldsymbol{\gamma}_i} \leftarrow \frac{\boldsymbol{\gamma}_i}{max(\boldsymbol{|\gamma}_i|)}, \boldsymbol{\Bar{\gamma}_i} = \frac{\sum_{d=1}^{D} \boldsymbol{\gamma_{i}^d}}{D}.
    \end{split}
\end{equation}
Here, the $| \cdot |$ denotes absolute value, and $d \in \{1, \cdots, D\}$ is a dimension index.  

\noindent An inverse scaling module, $\psi^{-1}$, is defined accordingly to re-scale back the predicted latent velocities, $\hat{\boldsymbol{\gamma}}_i$, before being passed to the decoder $\mathbf{D_\omega}$. The function of $\psi^{-1}$ can be formulated as
\begin{align}
    \notag \psi^{-1}({\hat{\boldsymbol{\gamma}}_i}) = \left[\hat{\boldsymbol{\gamma}_i} \times {\sqrt{\frac{\sum_{d=1}^{D} (\boldsymbol{\gamma_{i}^d} - \boldsymbol{\Bar{\gamma}}_{i})^2}{{D}}}} 
    + \boldsymbol{\Bar{\gamma}_i}\right] \times {max(\boldsymbol{|\gamma}_i|)}.
\end{align}

\noindent {\bf \em - Diffusion process with geometric conditioning.} Following the similar principles of~\cite{ho2020denoising}, we introduce a normal Gaussian noise, $\epsilon \sim \mathcal{N}(0, \mathbf{I})$, into the random noise generation process, with $\Sigma$ representing a covariance matrix. Given the initial latent velocity fields $\boldsymbol{\gamma}_{i,0}$, sampled from the real data distribution $q(\boldsymbol{\gamma})(\boldsymbol{\gamma}_{i,0} \sim q(\boldsymbol{\gamma}))$, we parameterize the forward diffusion process as a Markov chain with a number of $T$ steps. Specifically, at each step $\tau \in [1, \cdots,T]$ of the markov chain, we recursively add the defined Gaussian noise with a time-dependent variance  $\beta_\tau$. This produces a new latent variable with distribution $q(\boldsymbol{\gamma}_{i,\tau} | \boldsymbol{\gamma}_{i,\tau-1} )$. We mathematically formulate the forward diffusion process in the latent velocity space as follows: 
\begin{equation}
    q(\boldsymbol{\boldsymbol{\gamma}}_\tau \, | \, \boldsymbol{\gamma}_{i,\tau-1}) = \mathcal{N}(\boldsymbol{\gamma}_{i,\tau}; \sqrt{1-\beta_\tau}\boldsymbol{\gamma}_{i,\tau-1}, \beta_\tau \mathbf{I}).
    \label{eq:forward_diff}
\end{equation}

Following the work of~\cite{ho2020denoising}, we reformulate Eq.~\eqref{eq:forward_diff} as a single-step computation given the sampled noise and time step:
\begin{equation}
\label{eq:single_fwd_diff}
    \boldsymbol{\gamma}_{i,\tau} = \sqrt{\Bar{\alpha}_\tau}\boldsymbol{\gamma}_{i,0} + \sqrt{1-\Bar{\alpha}}_\tau\epsilon,  
\end{equation}
where $\Bar{\alpha}_\tau =\prod^{\tau}_{t=1}\alpha_t$ with $\alpha_\tau = 1 - \beta_\tau$.

In the reverse diffusion process, we first introduce a novel geometric conditioning in the latent velocity space. This enables our model to adapt more effectively to specific contexts; hence improving its predictive performance based on observed template images and provided text instructions. More specifically, our geometric conditioning is done by concatenating the deformation embedding - latent velocity fields, $\{\boldsymbol{\gamma}_{\tau,i}\}$, image embedding - downsampled template image gradient, $\{\nabla m_i\}$, and text embedding - encoded using a CLIP text encoder~\cite{radford2021learning}, $\{\Lambda_i\}$.

Our reverse diffusion process consists of sampling the denoised latent variable of velocity fields $\hat{\boldsymbol{\gamma}}_0$, starting from $\boldsymbol{\gamma}_{\tau, i}$ at the $\tau$-th time-step. The reverse process is given by 
\setlength{\abovedisplayskip}{3pt}
\begin{multline}
    p_\theta(\boldsymbol{\gamma}_{i,\tau-1} | \boldsymbol{\gamma}_{i,t}) = \mathcal{N}(\boldsymbol{\gamma}_{i,\tau-1}; \mu_\theta(\boldsymbol{\gamma}_{i,\tau}, \nabla m_i , \Lambda_i, \tau), \\ \Sigma_\theta(\boldsymbol{\gamma}_{i,\tau}, \nabla m_i , \Lambda_i, \tau)),
\end{multline}
where $\mu_\theta$ and $\Sigma_\theta$ represent the mean and variance for each time-step of the reverse process, such that $\Sigma_\theta(\gamma_{\tau}, \nabla m_i , \Lambda_i, \tau) = \sigma_\tau^2 \mathbf{I}$. This process removes the noises added to the latent velocity, which is practically implemented by a denoising network, $\mathbf{Z}_\theta$. In our experiments, we employ a UNet architecture~\cite{ronneberger2015unet} as a network backbone for ${\mathbf{Z}_\theta}$ and predict the noise to be removed. \\   

\noindent {\bf \em - Network loss.} The loss function of our proposed LCGD module that ensures optimal prediction of the noise to be removed in the latent velocity space is given as
\begin{equation}
\label{eq:lcgd}
\mathcal{L}_\theta = \frac{1}{N}\mathlarger{\sum}_{i=1}^N \vert\vert \epsilon - {\mathbf{Z}}_{\theta}
   \vert\vert^2_2 + \\
    \lambda \vert\vert \boldsymbol{\gamma}_{i,0} -\hat{\boldsymbol{\gamma}}_{i,0}  \vert\vert_2^2  + \text{reg}(\theta), 
\end{equation}
where $\text{reg}(\cdot)$ is the regularization applied on the network's parameters and $\lambda$ is a weighting parameter.

To achieve an optimal trade-off between sample quality and diversity of velocity fields in latent space post-training, we explore the best combination of conditional and unconditional diffusion models, guided by image-condition under guidance scale $\delta_\mathrm{I}$, the text-condition under guidance scale $\delta_\mathrm{T}$, and a null-condition $\emptyset$. Following similar principles in a recently developed work of classifier-free guidance~\cite{rombach2022highresolution}, we first jointly train a conditional and an unconditional TPIE model, and then combine the resulting score estimates to obtain the predicted noise $\hat{\mathbf{Z}}$ as
\setlength{\abovedisplayskip}{3pt}
\begin{multline}
\label{eq:cfg_train}  
\hat{\mathbf{Z}}_{\theta}(\gamma_{i,\tau}, \nabla m_i , \Lambda_i) \\= \mathbf{Z}_{\theta}(\boldsymbol{\gamma}_{i,\tau}, \emptyset, \emptyset) + 
 \delta_\mathrm{I} \cdot (\mathbf{Z}_{\theta}(\boldsymbol{\gamma}_{i,\tau}, \nabla m_i , \emptyset) -  \mathbf{Z}_{\theta}(\boldsymbol{\gamma}_{i,\tau}, \emptyset, \emptyset)) + 
 \\ \quad \quad \delta_\mathrm{T} \cdot (\mathbf{Z}_{\theta}(\boldsymbol{\gamma}_{i,\tau}, \nabla m_i , \Lambda_i) - \mathbf{Z}_{\theta}(\boldsymbol{\gamma}_{i,\tau}, \nabla m_i, \emptyset)),
 \\= (1-\delta_\mathrm{I}) \cdot \mathbf{Z}_{\theta}(\boldsymbol{\gamma}_{i,\tau}, \emptyset, \emptyset) + (\delta_\mathrm{I} - \delta_\mathrm{T}) \cdot (\mathbf{Z}_{\theta}(\boldsymbol{\gamma}_{i,\tau}, \nabla m_i , \emptyset) 
\\ \quad \quad + \delta_\mathrm{T} \cdot \mathbf{Z}_{\theta}(\boldsymbol{\gamma}_{i,\tau}, \nabla m_i , \Lambda_i).
\end{multline}

\subsection{Network Optimization}
The total loss of our TPIE framework includes the loss from both representation learning of velocity fields network (Eq.~\eqref{eq:registration}) and latent conditional geometric diffusion model (Eq.~\eqref{eq:lcgd}). Defining $r$ as the weighting parameter, we are now ready to write the joint network loss as $\mathcal{L} = \mathcal{L}_\omega + r \mathcal{L}_\theta$. We employ an alternative optimization scheme~\cite{nocedal1999numerical} to minimize the total loss. More specifically, we jointly optimize all network parameters by alternating between the training of the two subnetworks. 

A summary of our joint learning of TPIE through an alternating optimization is in Alg.~\ref{alg:training}. In testing phase, the process of sampling latent velocity fields from a randomly generated multivariate Gaussian distribution is in Alg.~\ref{alg:sampling}.
\begin{algorithm}[!htbp]
   \caption{TPIE Training}
   \label{alg:training}
    \begin{algorithmic}[1]
   \State \textbf{Input: } data $(m_i, f_i, \Lambda_i)_{i=1}^{N_{train}}$ where $N_{train}$ denotes the total number of pairwise images. 
   \Repeat
    \Repeat{} 
   \State Optimize the registration loss in Eq.~\ref{eq:registration}.
   \Until{converged} 
   \Repeat
   \State $\boldsymbol{\gamma}_{0,i} \gets \mathbf{E}_\omega$([$m_i, f_i$])
   \State Sample $\tau \sim Uniform(1, \mathrm{T})$, ~ $\epsilon \sim \mathcal{N}(0, \mathbf{I})$,  
   \State $\boldsymbol{\gamma}_{\tau,i} \gets q(\boldsymbol{\gamma}_{\tau, i} \vert \boldsymbol{\gamma}_{0, i})$
   \For{$t = \tau, ..., 1$}
   \State Get ${\boldsymbol{\gamma}}_{t-1,i}$ from $
   p_\theta(\boldsymbol{\gamma}_{t-1,i} \vert \boldsymbol{\gamma}_{t,i})$
   \EndFor
   \State Optimize the latent conditional geometric diffusion loss in Eq.~\ref{eq:lcgd}.
   \Until{converged}
   \State replace $\mathbf{E}_\omega([m_i, f_i]) \leftarrow \hat{\boldsymbol{\gamma}}_{0,i}$
   \Until{converged}
\end{algorithmic}
\end{algorithm}

\begin{algorithm}[!htbp]
   \caption{TPIE Sampling in Testing}
   \label{alg:sampling}
\begin{algorithmic}[1]
\State  \textbf{Input:} data $\{m_i, \Lambda_i\}_{i=1}^{N_{test}}$ 
\State Initialize $\boldsymbol{\gamma}_{T, i} \sim \mathcal{N}(0,\mathrm{I})$.
\For{$\tau = \mathrm{T},...,1$}
\State $\rho \sim \mathcal{N}(0,\mathbf{I})$ if $\tau>1$ else $\rho = 0$
\State $\boldsymbol{\gamma}_{\tau-1, i} = \frac{1}{\sqrt{\alpha_\tau}} \left(\boldsymbol{\gamma}_{\tau,i} - \frac{1-\alpha_\tau}{\sqrt{1-\Bar{\alpha_\tau}}} \hat{\mathbf{Z}}_{\theta}(\boldsymbol{\gamma}_{\tau,i}, \tau, \nabla m_i , \Lambda_i)\right) + \sigma_\tau\rho$
\EndFor
\State $\hat{f}_i = \mathbf{D_\omega}({\boldsymbol{\gamma}}_{0, i}, m_i)$
\State {\bfseries return} $\hat{f}_i$
\end{algorithmic}
\end{algorithm}

\section{Experimental Evaluation}

We validate the effectiveness of our proposed model, TPIE, on a diverse set of real-world image data that reflect geometric shape changes over time. \\

\noindent{\bf Komatsuna plant growth.} We include $300$ frames of RGB-D label-maps representing five Komatsuna plants from the publicly available Komatsuna data repository~\cite{Uchiyama2017komatsuna}. The label maps cover different leaves that emerge from the bud and grow in size over time, where the plant growth was monitored upto $228$-$236$ hours. All data frames were resampled to to $256^2$ and underwent pre-alignment with affine transformations.\\
\noindent{\bf 2D longitudinal brain MRI.} We include a total of $2618$ T1-weighted longitudinal (time-series) brain MRIs sourced from the Open Access Series of Imaging Studies (OASIS-3) dataset~\cite{LaMontagne2019oasis3}. This experiment aims to validate our method using longitudinal data that includes scans at varying time intervals for individuals spanning both healthy subjects and Alzheimer's diseases (AD), aged $60$-$90$. Given the scenario that many existing image generation/editing methods with text instructions focus on 2D natural images~\cite{brooks2023instructpix2pix, meng2021sdedit}, we specifically utilize 2D scans derived from this 3D brain data for comparison with state-of-the-art baselines. All MRIs undergo pre-processing, including resizing to $256^2$, with isotropic voxels of $1 \text{mm}^2$, skull-stripping, intensity normalization, bias field correction, and pre-alignment using affine transformations.\\
\noindent{\bf 3D shape of hippocampus.} We use $2618$ 3D hippocampi shapes of all subjects (cropped to the region of interest with a resolution of $64^3$) from the same OASIS-3 dataset. The pre-processing steps were kept identical to the brain MRIs.

\begin{figure*}[!htbp]
\centering
\includegraphics[width=\textwidth, trim = 0cm 0.1cm 0cm 0cm]{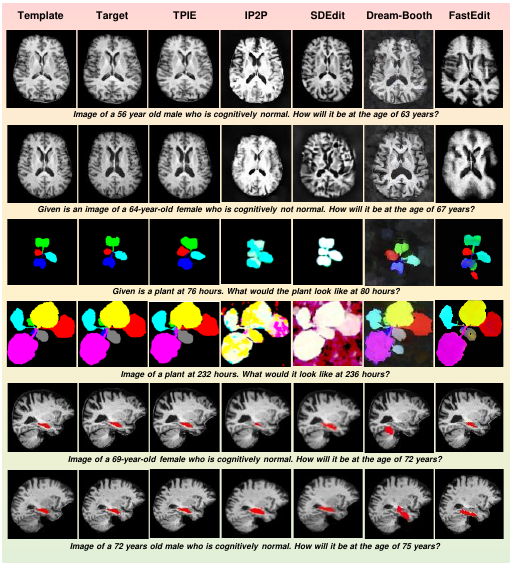}
\caption{A comparison of our method TPIE with all fine-tuned baselines. Left to right: given template images, target images, and generated samples. Top to bottom: examples of images generated from datasets of brain MRI, Komatsuna plants, and hippocampus shape. }
\label{main_preds}
\end{figure*}

\noindent{\bf Text instructions.} We formulate text instructions by initially describing the template image, followed by the description of the target image formatted as a question that we ask the model to generate. Each subject's information incorporates both clinical and biological variables, such as cognitive diagnosis, age, sex, and gender for brain and hippocampi data. The description of plant dataset includes the age of the plant in hours. We combine the images as pairs of template and target images that follow a pattern of incremental growth over time. Each template-target pair is associated with its corresponding text instruction. 

\subsection{Experiments and Implementation Details}
We evaluate the TPIE model from three perspectives: (i) the benefits of using geometric shape information in text-instructed latent conditional geometric diffusion model for realistic image generation, (ii) modeling the growth of geometric structures of objects presented in image, and (iii) predicting the trajectory of geometric changes over time. 

We compare our model with four primary baselines that offer publicly available, trainable source codes - \textit{(i) IP2P:} an image editing model based on conditional diffusion strategy \cite{brooks2023instructpix2pix}, \textit{(ii) SDEdit:} a guided image synthesis and editing model that uses score-based denoising diffusion approach through a stochastic
differential equation~\cite{meng2021sdedit}, \textit{(iii) DreamBooth:} a fine-tuned text-to-image model that mimics the appearance of subjects \cite{ruiz2022dreambooth} and \textit{(iv) FastEDit:} a fine-tuned text-to-image model that performs the edit based on semantic discrepancies  \cite{chen2024fasteditfasttextguidedsingleimage}. For a fair comparison, we fine-tune all baselines on the experimental datasets. \\

\noindent{\bf Evaluation on fidelity and relevance of generated images.} We first evaluate the model's capability to accurately learn the underlying latent distribution of geometric deformations (that reflect object topology) on all datasets. Three different performance metrics with varying number of input features, $(n)$, will be reported to extensively validate our method. These include the Fréchet Inception Distance (FID)~\cite{heusel2017gans} $(n=64)$, Kernel Inception Distance (KID)~\cite{binkowski2018demystifying} $(n=2048)$, and Inception Score (IS)~\cite{salimans2016improved}. Note that the first two metrics are computed between the features extracted from the real and generated images using a pre-trained inception model~\cite{szegedy2015rethinking}. The same model is used to obtain IS to validate the quality of generated images by computing the entropy of the predicted class labels. \\

\noindent{\bf Evaluation on reliability and certainty of model predictions.} We examine the reliability and confidence of our model in learning the pattern of growth and geometric shape changes over time. We compute the pixel-wise mean and standard deviation for all generated samples. For the plant dataset, we visualize the confidence intervals, which show the regions with $95\%$ confidence in growth patterns. The lower and upper bounds for this interval are obtained by calculating the mean of $70$ generated samples and computing pixel-wise bounds that are two standard deviations below and above the mean. For the brain and hippocampus datasets, we model the intermediate structural changes/progressions for a given pair of images that allows us to visualize the trajectory of changes over time. \\

\begin{figure*}[!htbp]
\centering
\centerline{\includegraphics[width=\textwidth, trim = 0cm 0.5cm 0cm 0cm]{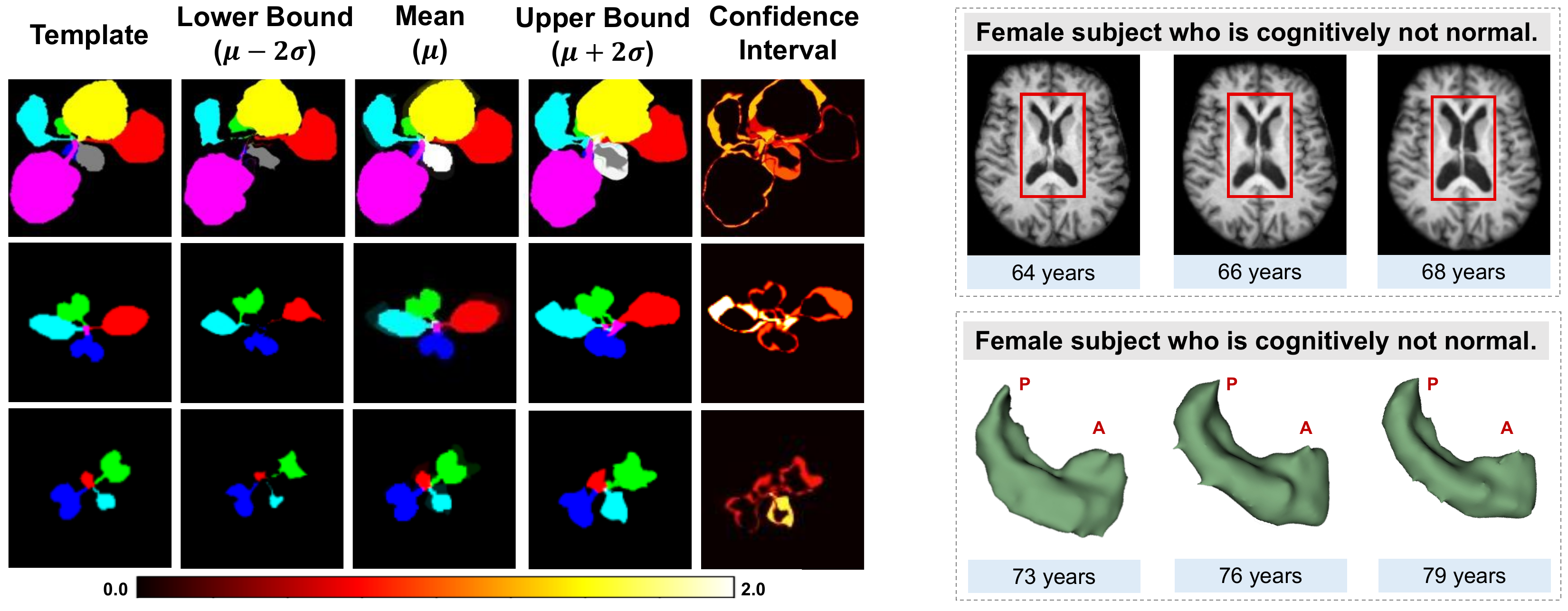}}
\caption{(a) A confidence map showing the lower/upper bound and confidence interval (regions with 95\% of ideal growth patterns) for plants generated by our model. (b) Examples of predicted time trajectory of growth patterns generated by our method. Top to bottom: axial view of brain MRIs (with brain ventricles outlined in red boxes); and 3D hippocampus shape with anterior (A) and posterior views (P).}
\label{plant_confidence}
\end{figure*} 

\noindent{\bf Parameter Setting.} We split all datasets into $80\%$, $10\%$, and $10\%$ for training, validation, and testing. We conduct all experiments on a server with one NVIDIA GeForce RTX 4090 and one NVIDIA A100 GPUs.  For the registration network, we use $LeakyReLU$ as activation function~\cite{maas2013rectifier}, while LCGD uses $SiLU$ activation~\cite{elfwing2017sigmoid}. All networks were trained with an initial learning rate of $1e^{-3}$ for the geometric deformation autoencoder network and $1e^{-5}$ for the LCGD model; with a decay of $1e-8$ after $1000$ steps. The variance schedule is fixed for the diffusion processes, ranging from $[0.0...1.0)$, and the maximum number of sampling steps is set to $500$. We train all the models using using the Adam optimizer~\cite{kingma2017adam} until convergence. 

\section{Results}
Fig.~\ref{main_preds} visualizes a comparison of the generated images (given a template image with conditional text) on all models. It shows that our model TPIE effectively preserves the topological structure of objects within the generated images. It also captures the progression of geometric changes over time (e.g., the expansion of brain ventricles with increased age). In contrast, samples generated by baselines fail to maintain the geometric integrity of various structures. For instance, in plant growth images, the leaves appear to merge or overlap unnaturally, disrupting their biological topology. Similarly, the baselines inaccurately predict parts of the hippocampus, generating regions that do not form the original structure of the brain.

\begin{table}[htbp]
\caption{A comparison of performance metrics over all methods for various datasets.}
\resizebox{\linewidth}{!}
{\begin{tabular}{clccc}
\toprule
                             &  $\text{Models}$          & $\text{FID}$ $\downarrow$  & $\text{KID}$ $\downarrow$           & $\text{IS}$   $\uparrow$        \\
                             \midrule
\multirow{5}{*}{\rotatebox{90}{$\text{Brain}$}}       & $\text{IP2P}$  & $4.758$ & $0.34(1.72e-7)$  & $1.21(0.23)$    \\
                             & $\text{SDEdit}$     & $5.395$ & $0.43(2.34e-7)$  & $1.62(0.28)$    \\
                             & $\text{DreamBooth}$ & $53.47$ & $0.77(3.81e-6)$  & $1.00(0.00)$    \\
                             & $\text{FastEdit}$   &   $3.789$    &      $0.22(1.63e-7)$          &    $1.36(0.19)$          \\
                             & $\text{TPIE}$       & $\mathbf{0.410}$ & $\mathbf{0.20(1.95e-5)}$  & $\mathbf{1.65(0.15)}$    \\
                             \midrule
\multirow{5}{*}{\rotatebox{90}{$\text{Plant}$}}       & $\text{IP2P}$       & $16.69$ & $0.27(1.99e-7)$  & $1.41(0.30)$    \\
                             & $\text{SDEdit}$     & $31.45$ & $0.20(6.39e-6)$  & $1.40(0.32)$   \\
                             & $\text{DreamBooth}$ & $83.77$ & $0.29(1.46e-5)$  & $1.00(0.00)$    \\
                             & $\text{FastEdit}$   &  $66.23$     &  $0.52(0.01)$              &      $1.13(0.07)$         \\
                             & $\text{TPIE}$       & $\mathbf{0.128}$  & $\mathbf{0.12(1.88e-7)}$ & $\mathbf{1.42(0.20)}$   \\
                             \midrule
\multirow{5}{*}{\rotatebox{90}{$\text{Hippocampus}$}} & $\text{IP2P}$       & $0.127$  & $0.03(1.22e-5)$ & $1.09(0.10)$    \\
                             & $\text{SDEdit}$     & $0.010$  & $0.30(2.16e-7)$  & $1.39(0.09)$    \\
                             & $\text{DreamBooth}$ & $86.25$ & $0.61(1.11e-7)$  & $1.00(8.43e-8)$ \\
                             & $\text{FastEdit}$   &  $4.497$     &     $ 0.09(2.17e-7)$       &       $\mathbf{1.60(0.34)}$        \\
                             & $\text{TPIE}$       & $\mathbf{0.005}$ & $\mathbf{0.02(1.79e-7)}$ & $1.25(0.04)$ \\
                             \bottomrule
\end{tabular}}
\label{table_comp}
\end{table}

Tab.~\ref{table_comp} reports the fidelity and relevance metrics computed between the real and generated images for TPIE and all the baselines with significantly lower FID and KID scores. Moreover, all models, including TPIE, achieve similar inception scores with small variance, providing a relative comparison of similarity scores where our method demonstrates better/comparative performance. 

The left panel of Fig.~\ref{plant_confidence} illustrates exemplary confidence maps of plant growth images generated from our model TPIE. It suggests that TPIE effectively captures the growth pattern over time, primarily focusing around the boundaries of the leaves. The right panel of Fig.~\ref{plant_confidence} displays a generated sequence of hippocampus along time. It shows that the pattern of brain ventricles increase in size (top) vs. anterior hippocampus shrinks (bottom) over time. This indicates that our model TPIE is able to learn the geometric changing process from the latent distribution of velocity fields. \\
\noindent {\bf Discussions \& Limitations.} One additional advantage of our model is that it can be used for several downstream tasks, such as topology-preserved data augmentation, classification and time series analysis. We validate our results via a binary classification task for brains (AD vs. healthy) by training a convolutional neural network~\cite{atlas1987artificial} and find that the images generated by our method yields an accuracy of $85\%$, while those generated by IP2P~\cite{brooks2023instructpix2pix} result in merely $49\%$ accuracy. This initial analysis demonstrates a promising performance of our model, TPIE. More results are presented in the supplementary materials. An additional advantage of TPIE is its ability to quantitatively assess topological changes in objects generated by diffusion models.

Our proposed model currently relies on training images with consistent object topology. Future work will be investigated to integrate variations of image  intensity and textures. This will further extend the applicability of the method to a wider variety of real-world image datasets, such as the waterbirds dataset from WILDs~\cite{koh2021wilds}.

\section{Conclusion}
This paper presents a novel method, named {\em Diffeomorphic Image Generation with Instructions from Text (TPIE)}, that for the first time incorporates the topology and geometry of objects in synthesized images using text-guided generative diffusion models. The newly developed latent conditional geometric diffusion module is trained to generate random object deformations, conditioned on the input template image and text instructions. Our experimental findings indicate that TPIE surpasses state-of-the-art models in generating images with well-preserved topological structure of objects. It's worth noting that our current model mainly focuses on the applicability of maintaining consistent topological characteristics across template-target training images. Future work will further develop a comprehensive and advanced model that offers flexibility for image generations within integrated spaces, encompassing both image intensity/texture and geometric shape. \\

\noindent {\bf Acknowledgment.}\\
This work was supported by NSF CAREER Grant 2239977.
{
    \small
    \bibliographystyle{ieeenat_fullname}
    \bibliography{main}
}

\end{document}